\documentclass[sigconf,nonacm=true]{acmart}
\usepackage[utf8]{inputenc}
\usepackage[encapsulated]{CJK}

\usepackage{booktabs}
\usepackage{adjustbox}
\usepackage{multirow}
\usepackage{booktabs}

\AtBeginDocument{%
  \providecommand\BibTeX{{%
    \normalfont B\kern-0.5em{\scshape i\kern-0.25em b}\kern-0.8em\TeX}}}

\input{zhwinfonts}
\begin{document}

\title{Response to LiveBot: Generating Live Video Comments Based on Visual and Textual Contexts}

\author{Hao Wu}
\affiliation{%
  \institution{ADAPT Centre,
School of Engineering, Trinity College Dublin}
  \city{Dublin}
  \country{Ireland}}
\email{hao.wu@adaptcentre.ie}

\author{Gareth J. F. Jones}
\affiliation{%
  \institution{ADAPT Centre, School of Computing, Dublin City University}
  \city{Dublin}
  \country{Ireland}
}
\email{Gareth.Jones@dcu.ie}

\author{Fran\c{c}ois Piti\'e}
\affiliation{%
 \institution{ADAPT Centre,
School of Engineering, Trinity College Dublin}
 \city{Dublin}
  \country{Ireland}}
\email{pitief@tcd.ie}

\begin{abstract}

Live video commenting systems are an emerging feature of online video sites. Recently the Chinese video sharing platform Bilibili, has popularised a novel captioning system where user comments (``\textit{danmu}'') are displayed as streams of moving subtitles overlaid on the video playback screen and broadcast to all viewers in real-time.

LiveBot was recently introduced as a novel Automatic Live Video Commenting (ALVC) application.
This enables the automatic generation of live video comments from both the existing video stream and existing viewers comments.
In seeking to reproduce the baseline results reported in the original Livebot paper, we found differences between the reproduced results using the project codebase and the numbers reported in the paper. 

Further examination of this situation suggests that this may 
be caused by a number of small issues in the project code, including a non-obvious overlap between the training and test sets. In this paper, we study these discrepancies in detail and propose an alternative baseline implementation as a reference for other researchers in this field.

\end{abstract}


\maketitle

\section{Introduction}
\begin{figure}[t]
  \centering
  \includegraphics[width=\linewidth]{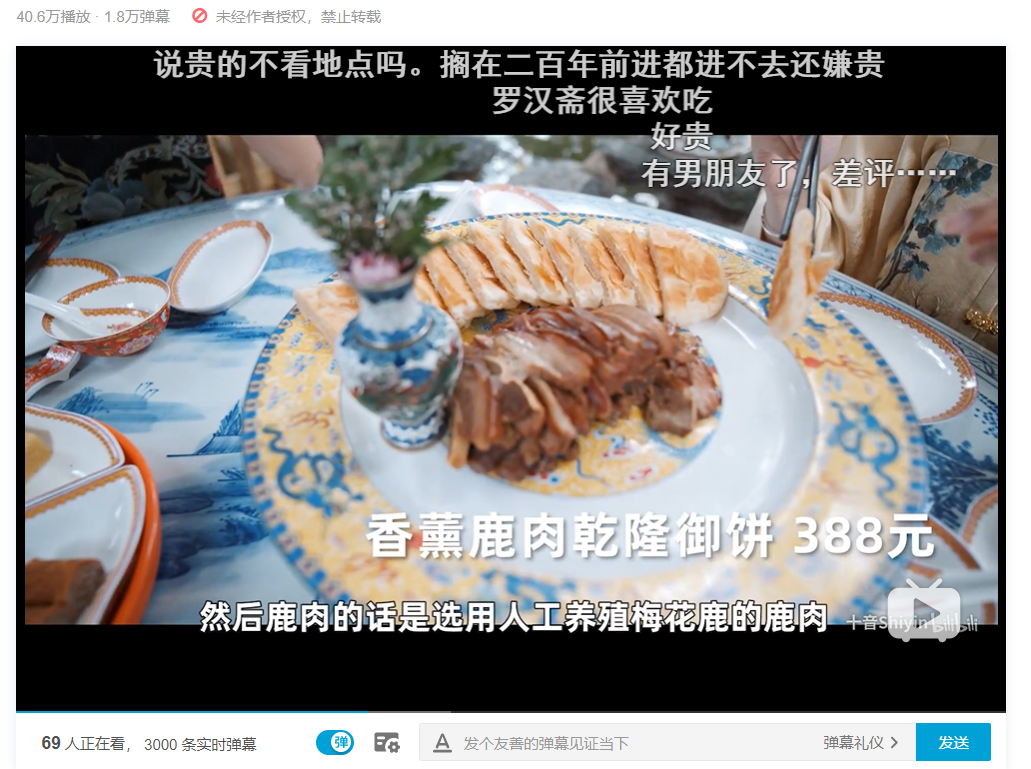}
  \caption{Screen shot of the video interface from bilibili.tv where danmu comment are floating though the top side of the video.}
  \label{fig:normal_video}
\end{figure}

Live commenting mechanisms has become a core feature for video platforms like Bilibili, one of the most popular video sharing platform in China with more than 300 millions monthly active users~\footnote{\hyperlink{http://ir.bilibili.com/news-releases/news-release-details/\-bilibili-inc-announces-first-quarter-2019-financial-results/}{http://ir.bilibili.com/news-releases/news-release-details/bilibili-inc-announces-first-quarter-2019-financial-results/}}. This feature increases user interaction by providing a real-time commentary subtitle system that displays user comments as streams of moving subtitles overlaid on the video playback screen, visually resembling a danmaku shooter game. These comments are are simultaneously broadcast to all viewers in real-time, they were originally called ``\textit{danmaku}'' in the Nicovideo Japanese platform and then 
\begin{CJK}{UTF8}{gbsn}
``弹幕'' 
\end{CJK}
in the Chinese Bilibili platform. In the rest of this document we refer to them by the Pinyin (a romanization system for Chinese characters) version: ``danmu''. Figure~\ref{fig:normal_video} shows an example of a video from Bilibili with a few danmu overlaid. The danmu system is different from the commenting system or online streaming system in most of the video sharing platforms, since it provides a chat room experience in which users can watch and discuss together. 

Several efforts have been made to investigate this new type of media content~\cite{lv2019gossiping,li2019visual,bai2019stories}. The creation of new danmu comments to enrich videos has the potential to improve the viewing experience of viewers and to help attract more viewers.
Shuming et al.~\cite{Livebot} proposed ``Livebot'', which use a unified transformer architecture to automatically generate new danmu comments given existing danmu comments and video frames. 
%
LiveBot uses AI agents to comprehend the videos and to interact with human viewers who also make 
comments. In this work, 
a large-scale live comment dataset with 2,361 videos and 895,929 live comments was constructed.
%
In an attempt to replicate the proposed method~\cite{Livebot}, using this dataset whih was provided to the research community by the authors, we found that our results were much lower than the reported baselines. To understand
this issue we carefully reviewed the implementation and dataset provided by the authors on their project webpage, and found a number of potential issues which may explain this discrepancy. We examine these problems one by one and analyse their impacts. Finally we propose a new baseline implementation which could serve as an independent reference.

\section{Background}

In this section we briefly introduce the background of ALVC task. We refer readers to the original paper~\cite{Livebot} for more detailed information.

\subsection{Task Formation and dataset}

The live commenting dataset 
built 
for the Automatic Live Video Commenting (ALVC) task was
collected from Bilibili and contains 2361 videos and 895,929 comments. Each video comment is associated 
with its related video time tag, which indicates where in the video each comment should appear. The processed dataset partition, the raw dataset and the code are available
at the GitHub page~\cite{livebot_github}.
In this paper we use this dataset for our experiments,
Table~\ref{table:dataset} shows the detailed statistics of the dataset.

\begin{table}[t]
\caption{Statistics on the Livebot dataset.}
\label{table:dataset}
\begin{tabular}{|l|r|r|r|r|}
\hline
Statistic            & Train     & Dev     & Test    & Total     \\ \hline
\#Video              & 2,161     & 100     & 100     & 2,361     \\ \hline
\#Comment            & 818,905   & 42,405  & 34,609  & 895,929   \\ \hline
\#Word               & 4,418,601 & 248,399 & 193,246 & 4,860,246 \\ \hline
Avg. Words           & 5.39      & 5.85    & 5.58    & 5.42      \\ \hline
Total Duration (hrs) & 103.81    & 5.02    & 5.01    & 113.84    \\ \hline
\end{tabular}
\end{table}
\begin{table}[t]
\caption{Statistics of new dataset.}
\label{table:newdata}
\begin{tabular}{|l|r|r|r|r|}
\hline
Statistic            & Train     & Dev     & Test    & Total     \\ \hline
\#Video              & 2,122     & 100     & 100     & 2,322     \\ \hline
\#Comment            & 788,645    & 34,581   & 34,767   & 857,993   \\ \hline
\#Word               & 4,083,379 & 178,172 & 176,115 & 4,437,666 \\ \hline
Avg. Words           & 5.17      & 5.15    & 5.06    & 5.17      \\ \hline
\end{tabular}
\end{table}

The ALVC task is defined as follows: given a video $\textbf{V}$, a time-stamp $\textit{t}$ and the surrounding comments $\textbf{C}$ near the time-stamp, the commenting system
should
generate a comment $\textbf{Y}$ relevant to the clips and/or the other comments near the time-stamp. 
Specifically, the model takes the nearest m frames ($I = {I_1, I_2, \dots, I_m}$) and $n$ comments ($C = {C_1, C_2, \dots, C_n}$) from the time-stamp $\textit{t}$ as input, and aims to generate
a
comment y = {$y_1, y_2, \dots, y_k$}.

\subsection{Model Structure}

\begin{figure}[t]
    \centering
    \includegraphics[width=\linewidth]{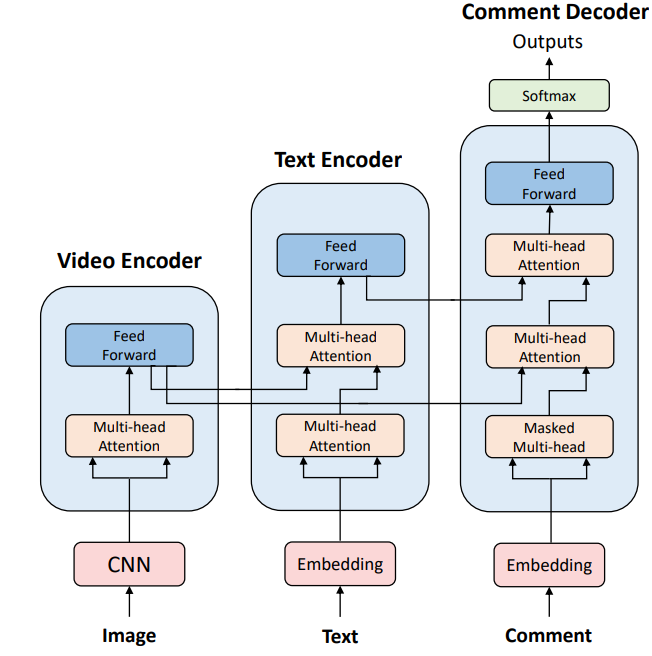}
    \caption{The structure of the unified transformer model, the encoding CNN is a pre-trained resnet18.}
    \label{Livebot_model}
\end{figure}

For our investigation 
we follow the model structure 
described in ~\cite{Livebot}, and illustrated 
in Figure~\ref{Livebot_model} 
(see section ``Model II: Unified Transformer Model'' in~\cite{Livebot}). 
Comments and video frames are encoded using a
Transformer architecture~\cite{vaswani2017attention}. The model  consists of 3 parts: the video encoder which encodes video frames into a visual representation; the text encoder which generates the contextual vector by encoding the sequence of input words combined with
the visual representation; and finally 
the comment encodes which combines these vectors in comment decoder to generate output tokens recursively.

\subsection{Evaluation Metrics}
\label{evaluation}

Retrieval based evaluation metrics are used in the reported experiments to
automatically evaluate ALVC: a candidate comment set is constructed for each test sample, then the model is asked to sort the candidate set;
the authors assume that a good model is able to rank the correct comments at the top
of the set proposed comments.

The candidate set contains 4 types of comments:
\begin{itemize}
    \item \textbf{Correct}: 5 groundtruth comments from humans.
    \item \textbf{Plausible}: 30 comments most similar to the title of the video based on tf-idf score.
    \item \textbf{Popular}: 20 comments 
    in the training set.
    \item \textbf{Random}: 
    Random comments taken from the training set to ensure there are 100 unique comments in the generated output set.
\end{itemize}

The following retrieval metrics are 
used to
evaluate the results: 

\begin{itemize}
\item \textbf{Recall@k}: the proportion of human comments found in the top-k recommendations, 
\item \textbf{Mean} Rank: the mean rank of the human comments, 
\item \textbf{Mean Reciprocal Rank}: the mean reciprocal rank of the human comments.
\end{itemize}

Results for all these metrics are presented in Table~\ref{table:summary_result}. We also report the confidence interval for each of these metrics. For recall@k we use the confidence interval for population proportions with confidence level at 95\% and for MR and MRR, we use the confidence interval with same confidence level. 

\section{Identified Issues in Reproducing Livebot Results}

We first tried to reproduce the work of~\cite {Livebot} using
the released the code and the dataset, 
For reference, the Livebot results are reported in Table~\ref{table:summary_result} and labeled with ``Livebot paper''. Specifically, results with different input are reported (e.g. ``Text Only'' means text input are all masked during test stage).  
We conducted our experiments using the code provided on the authors'
Github project page, and used
the same model structure and configurations (batch size, learning rate etc.) described in~\cite{Livebot}. The results we obtained
are shown in the same table with the label ``Issue \#1''.
Clearly the results from our experiments are
much lower than the baselines. In order to explore the reasons for
the performance mismatch, we conduct a series of investigations examining
the GitHub implementation and the released dataset.
From our investigation, we have identified a number of issues with GitHub implementation 
which are presented below.

\subsection{Issue \#1: Candidate Set Ranking.}

First, in the implementation, the re-ranked candidate list is sorted based on the cross-entropy loss in descending order. However according to the paper, a good candidate should be placed at the topside of the candidate list, in this case the cross-entropy loss should be sorted in ascending order. 
This issue is also raised in the GitHub issue page by another researcher~\footnote{https://github.com/lancopku/Livebot/issues/1}, the corresponding results are
labeled 
`GitHub Issue'' in Table~\ref{table:summary_result}. 
We report the results with this issue fixed (see ``Issue \#1-2''). The scores are very close to the results from the GitHub issue page, we can see that after fixing the ranking problem the scores improve a little, but are still significantly lower than the reported Livebot baselines.

\subsection{Issue \#2: Candidate Scores}

We then carefully looked at the evaluation code and noticed a subtle error in the candidate score computing: in the original implementation the score of a candidate is computed as the sum of the cross-entropy loss for every token rather than the mean value. This 
results in an advantage for short candidates and, in fact, we found that the top re-ranked positions in the list are mostly occupied by comments of only one word.

We fixed the code by averaging the score over every non-ignored token (tokens for padding and separating are ignored when computing cross-entropy loss). Thus
instead of
\begin{equation}
    \mathrm{Score}(c) = \sum_{i=0}^{L}\mathrm{Cross Entropy}(g_{i},h{i})\,,
\end{equation}
we implemented:
\begin{equation}
    \mathrm{Score}(c) = \frac{\sum_{i=0}^{L}\mathrm{Cross Entropy}(g_{i},h_{i})}{\mathrm{\#Valids}}\,,
\end{equation}
where $g_{i}$ and $h_{i}$ are the i-th output token and ground truthtoken, $L$ is the maximum length of the model output (including padding), $\mathrm{\#Valids}$ is the number of valid tokens in a candidate. The results reported as ``Issue \#3'' in Table~\ref{table:summary_result}, at this step we obtain
scores that are closer to the baselines.

\subsection{Issue \#3: Construction of the Plausible Set.} 

We also found an
inconsistency in constructing the plausible set. It is described in the paper that when building the candidate list, the plausible set is retrieved based on the video title. However, in the implementation we noticed that the plausible set is retrieved using current context comments (The comments surrounding 
the ground truth comment, which is also the text input) as the query rather than video title. 
Unfortunately, the mapping between the raw dataset and the provided dataset are not given, so we are not able to reconstruct the provided dataset from the raw dataset, and hence could not direct compare the results with and without fixing this issue. In our final
experiments (``Issue \#1-4'') reported in the next section, we 
follow the Livebot paper and use video title to construct plausible set.
 
\begin{table*}[t]
\caption{Results of conducted experiments, ``Run Label'' section shows the version of the experiment (e.g. Issue \#1-3 means the experiments with issue 3 and all past issues fixed), Recall@k, MRR: higher is better; MR: lower is better.}
\label{table:summary_result}
\centering
\begin{tabular}{|l|l|l|l|l|l|l|l|l|}
\hline
Input                        & Model          & Dataset           & Run Label        & Recall@1 & Recall@5 & Recall@10 & MR & MRR \\ \hline
\multirow{6}{*}{Text Only}   & Livebot        & unknown               & Livebot paper   &13.95               & 34.57              & 51.57                & 17.01            & 0.251           \\ 
 & Livebot        & Provided          & Issue \#1           & 5.41  $\pm$  0.05    & 20.33 $\pm$ 0.18   & 34.58 $\pm$ 0.31    & 23.78 $\pm$ 0.44 & 0.147 $\pm$ 0.01  \\ 
 & Livebot        & Provided          & Issue \#1-2         & 9.77  $\pm$ 0.87    & 24.31 $\pm$ 0.22   & 31.15 $\pm$ 0.28    & 21.10 $\pm$ 0.51 & 0.185 $\pm$ 0.008 \\ 
 & Livebot        & Provided          & Issue \#1-3         & 17.11 $\pm$ 0.16    & 37.07 $\pm$ 0.35   & 51.08 $\pm$ 0.48    & 14.91 $\pm$ 0.48 & 0.280 $\pm$ 0.009 \\ 
 & Livebot        & No duplicate      & Issue \#1-4         & 12.04 $\pm$ 0.11    & 25.01 $\pm$ 0.23   & 42.04 $\pm$ 0.40    & 20.77 $\pm$ 0.65 & 0.219 $\pm$ 0.01  \\  
 & OpenNMT        & No duplicate      & Re-Implementation   & 12.48 $\pm$ 0.12    & 24.16 $\pm$ 0.22   & 42.68 $\pm$ 0.41    & 18.66 $\pm$ 0.49 & 0.228 $\pm$ 0.01  \\ \hline
                             
\multirow{6}{*}{Visual Only} & Livebot        & unknown               & Livebot paper   & 11.40    & 32.62   & 50.47     & 18.12 & 0.231\\ 
                             & Livebot        & Provided          & Issue \#1           & 5.44 $\pm$ 0.05    & 20.30 $\pm$ 0.19   & 36.31 $\pm$ 0.35     & 23.84 $\pm$ 0.45 & 0.142 $\pm$ 0.01    \\ 
                             & Livebot        & Provided          & Issue \#1-2         & 8.66 $\pm$ 0.08    & 22.64 $\pm$ 0.21   & 31.42 $\pm$ 0.30     & 21.23 $\pm$ 0.54 & 0.175 $\pm$ 0.009   \\ 
                             & Livebot        & Provided          & Issue \#1-3         & 7.78 $\pm$ 0.07    & 26.78 $\pm$ 0.25   & 40.23 $\pm$ 0.38     & 19.66 $\pm$ 0.54 & 0.183 $\pm$ 0.01    \\ 
                             & Livebot        & No duplicate      & Issue \#1-4         & 6.55 $\pm$ 0.06    & 23.41 $\pm$ 0.22   & 39.38 $\pm$ 0.38     & 20.77 $\pm$ 0.67 & 0.169 $\pm$ 0.01    \\  
                             & OpenNMT        & No duplicate      & Re-Implementation   & 7.01 $\pm$ 0.06    & 24.35 $\pm$ 0.23   & 37.76 $\pm$ 0.36     & 19.89 $\pm$ 0.43 & 0.172 $\pm$ 0.01    \\ \hline
                             
\multirow{7}{*}{Text+Visual} & Livebot        & unknown           & Livebot paper       &18.01    & 38.12    & 55.78     & 16.01 & 0.275\\ 
                             & Livebot        & Provided          & Issue \#1           & 5.81 $\pm$ 0.05    & 21.49 $\pm$ 0.19   & 36.43 $\pm$ 0.35     & 22.22 $\pm$ 0.45 & 0.155 $\pm$ 0.01\\ 
                             & Livebot        & Provided          & Issue \#1-2         & 11.46 $\pm$ 0.11   & 26.22 $\pm$ 0.24   & 32.96 $\pm$ 0.29     & 19.54 $\pm$ 0.48 & 0.204 $\pm$ 0.009\\ 
                             & Livebot        & Provided          & GitHub Issue        & 10.56              & 25.24           & 34.05                & 20.26            & 0.170\\ 
                             & Livebot        & Provided          & Issue \#1-3         & 18.79 $\pm$ 0.17   & 39.46 $\pm$ 0.38   & 50.13 $\pm$ 0.48     & 16.17 $\pm$ 0.46 & 0.297 $\pm$ 0.01\\ 
                             & Livebot        & No duplicate      & Issue \#1-4         & 15.50 $\pm$ 0.14   & 34.57 $\pm$ 0.33   & 48.48 $\pm$ 0.46     & 17.25 $\pm$ 0.48 & 0.260 $\pm$ 0.01\\ 
                             & OpenNMT        & No duplicate      & Re-Implementation   & 14.79 $\pm$ 0.14   & 33.45 $\pm$ 0.32   & 48.93 $\pm$ 0.46     & 17.45 $\pm$ 0.49 & 0.257 $\pm$ 0.01    \\ \hline
                             
\end{tabular}
\end{table*}

\subsection{Issue \#4: the Training/Testing Set}
\label{duplicate}

We carefully examined the released dataset, specifically we checked the overlapped comments across the training and test set of the given processed dataset. There are 5,436 out of 17,771 comments in the test set also appear in the training set. Although some popular comments can be expected to 
appear in different videos, after manually checking the provided dataset we found that there are a number of identical videos assigned with different video ids that appear in both the training and test sets. Table~\ref{table:identical_video} lists several examples we found of this situation. In the raw dataset we use video title to uniquely identify a video and found that there are 38 videos which appeared more than once in the raw dataset.  

To address this issue, we decide to build the dataset from the raw dataset rather than directly update the processed dataset due to the lack of video mapping between the raw dataset and the processed dataset. After removing redundant videos from the raw dataset we end up with 2322 unique videos. We follow the Livebot paper and split the training / development / test set into 2,122 / 100 / 100 videos and conducted experiments with all above issues fixed. (statistics of the dataset are summarised in Table~\ref{table:newdata}). This dataset is labeled as ``No duplicate'' in the result table.

Our results after removing the duplicate videos are shown
as  ``Issue \#1-4'' in Table~\ref{table:summary_result}. Compared to ``Issue \#1-3'' the performance can be observed to be slightly
lower, 
which is what we 
could anticipate since the model no longer gains from the overlapped information across the training and test set.
\begin{CJK}{UTF8}{gbsn}
\begin{table*}[t]
\caption{Several Comments that appear both in training and test data set of the provided dataset.}
\label{table:identical_video}
\centering
\begin{tabular}{|l|l|}
\hline
Comments & Translation       \\ \hline
\begin{tabular}[c]{@{}l@{}}像我这么瘦的可能效果不会太明显，各种无器械动作交杂着\\ 做两个多月才有了明显的变化，还不是很大\end{tabular} & \begin{tabular}[c]{@{}l@{}}It might not be obvious for skinny people like me,\\  there are only minor changes after 2 month of exercise.\end{tabular}    \\ \hline
\begin{tabular}[c]{@{}l@{}}这样看不出，第一和最后一天再对比下，长肌肉不是做了多\\ 少，是休息恢复，增长多少，不建议天天练\end{tabular} & \begin{tabular}[c]{@{}l@{}}Can not tell anything from this, muscle growth is about\\  resting and recovering rather than work out everyday.\end{tabular} \\ \hline
100个考验耐力，到后期，强度就不高了，复合俯卧撑最好 & \begin{tabular}[c]{@{}l@{}}Doing 100 requires endurance not strength, compound\\  push-up is the best.\end{tabular}                                      \\ \hline
\begin{tabular}[c]{@{}l@{}}每天100个俯卧撑100个仰卧起坐跑步10公里坚持3年然后再\\ 把头发剃光滑稽\end{tabular}         & \begin{tabular}[c]{@{}l@{}}100 push-ups 100 sit-ups 100 squats and a 10km run \\ every single day for 3 years then shave your hair lol.\end{tabular}     \\ \hline
\begin{tabular}[c]{@{}l@{}}练肌肉最费钱，想练快就每天吃低脂牛肉，配合锻炼，半\\ 年就有显著变化\end{tabular}              & \begin{tabular}[c]{@{}l@{}}Muscle gain is expensive, regular exercise with \\ low-fat beef and you will see the changes in half a year\end{tabular}      \\ \hline
\end{tabular}
\end{table*}

\section{Re-Implementation Using OpenNMT}
\end{CJK}
In order to provide a reproducible implementation for later research on the ALVC task, we re-implemented the transformer network of LiveBot using the OpenNMT~\cite{klein-etal-2017-opennmt} open-source neural machine translation framework. 
We followed the model structure shown in figure~\ref{Livebot_model}, and used the newly constructed dataset described in section~\ref{duplicate}, with all duplicate videos removed.

The vocabulary size is set to 30,000 to keep it consistent with the original paper, and in the transformer network, the size of the word embedding and hidden layer are set to 512, as in~\cite{Livebot}. Additionally, the batch size is set to 64 and dropout rate to 0.2. The optimization method is chosen as Adam~\cite{kingma2014adam}, with $\beta_{1}=0.9$ and $\beta_{2}=0.998$.
Results of this re-implementation are reported in Table~\ref{table:summary_result} under the label ``Re-implementation''. With past issues resolved. At this stage the scores we are get are very close to run ``Issue \#1-4'', we believe the implementation and scores generated are valid and could serve as a new baseline for this task.

The code and the dataset used to generate the above result is available on GitHub~\footnote{https://github.com/fireflyHunter/OpenNMT-Livebot}

\section{Conclusions}

In this paper we reviewed the code presented as the official LiveBot implementation and found a number of discrepancies with the original paper. We have addressed each of these issues and reported updated results accordingly. We also propose a new baseline implementation using the OpenNMT framework. The updated baseline results are still lower than the ones reported in the original Livebot paper. However, since we do not access to the exact version of code used to produce the results these original results are are not able to determine the exact reason for these differences,
but based on our experiments and out analysis, we believe this performance gap is caused by the removal of the duplicate videos.  

\section*{Acknowledgement}

This work was supported by Science Foundation Ireland as part of the ADAPT Centre (Grant 13/RC/2106) (\url{www.adaptcentre.ie}) at Trinity College Dublin.

\bibliographystyle{ACM-Reference-Format}
\bibliography{sample-base}

\end{document}